%% file: emnlp2023.tex
\pdfoutput=1

\documentclass[11pt]{article}

\usepackage[]{EMNLP2023}

\usepackage{times}
\usepackage{latexsym}

\usepackage[T1]{fontenc}

\usepackage[utf8]{inputenc}

\usepackage{microtype}

\usepackage{inconsolata}

\usepackage{url}

\usepackage{xcolor}

\usepackage{multirow}
\usepackage{graphicx}
\usepackage{subfig}
\usepackage{amsfonts}
\usepackage{bm}

\usepackage{amsmath,array}

\usepackage{booktabs}

\usepackage{pifont}
\usepackage{colortbl}
\usepackage{xcolor}

\usepackage{quoting}
\quotingsetup{vskip=0pt}

\title{AHP-Powered LLM Reasoning for Multi-Criteria Evaluation of Open-Ended Responses}

\author{Xiaotian Lu\thanks{~~Corresponding author. }$~\hspace{0.5mm}^1$ Jiyi Li $^2$  Koh Takeuchi $^1$ Hisashi Kashima $^1$ \vspace{1mm}\\
$^1$Kyoto University \hspace{5mm}$^2$University of Yamanashi\\
{\small $^1$ Department of Intelligence Science and Technology, Graduate School of Informatics, Kyoto University, Kyoto, Japan } \\
{\small $^1$\texttt{{\{lu@ml.ist.,takeuchi@,kashima@\}}i.kyoto-u.ac.jp}}\\
{\small $^2$ Department of Computer Science and Engineering, University of Yamanashi, Kofu, Japan } \\
{\small $^2$\texttt{jyli@yamanashi.ac.jp}} \\
}

\begin{document}
\maketitle
\begin{abstract}
Question answering (QA) tasks have been extensively studied in the field of natural language processing (NLP). 
Answers to open-ended questions are highly diverse and difficult to quantify, and cannot be simply evaluated as correct or incorrect, unlike close-ended questions with definitive answers. 
While large language models (LLMs) have demonstrated strong capabilities across various tasks, they exhibit relatively weaker performance in evaluating answers to open-ended questions. In this study, we propose a method that leverages LLMs and the analytic hierarchy process (AHP) to assess answers to open-ended questions. We utilized LLMs to generate multiple evaluation criteria for a question. Subsequently, answers were subjected to pairwise comparisons under each criterion with LLMs, and scores for each answer were calculated in the AHP. We conducted experiments on four datasets using both ChatGPT-3.5-turbo and GPT-4. Our results indicate that our approach more closely aligns with human judgment compared to the four baselines. Additionally, we explored the impact of the number of criteria, variations in models, and differences in datasets on the results.
\end{abstract}

\input{intro.tex}

\input{related}

\input{proposed}

\input{results}
\input{conclusion}

\section*{Limitations}
This study involved experiments across two task types, four datasets, and two large models. Due to time and financial constraints, our experiments do not extend to more tasks currently. Compared to simpler baseline methods, our approach requires additional computations within LLMs, which in turn increases both time and financial costs.

\section*{Acknowledgements and Ethics Statement}
This work was supported by JST CREST Grant Number JPMJCR21D1, JSPS KAKENHI Grant Number JP23KJ1215 and JSPS KAKENHI Grant Number JP23K28092.

We have abided by the ACL Ethics Policy in this work. No personal information or data was used in this study.

\bibliography{anthology,custom}
\bibliographystyle{acl_natbib}




\end{document}

%% file: intro.tex
\section{Introduction}

Large language models (LLMs) have shown remarkable capabilities in a wide range of tasks, such as natural language generation~\cite{karanikolas2023large,ko2024natural}, summarization~\cite{ahmed2022few,laban2023summedits}, translation~\cite{huang2023towards} and text classification~\cite{moller2024parrot}. Question Answering (QA) tasks have been widely studied and can be utilized to assess the breadth of knowledge and logical comprehension abilities of LLMs. These tasks are designed to probe the models not only for their factual accuracy but also for their ability to infer answers from complex queries, thereby reflecting their understanding of both context and content.

Most studies on QA tasks focus on {\it close-ended} questions~\cite{tan2023can, han2023robustqa,zhu2021retrieving}, which have definitive correct answers. For example, the following are three close-ended questions~\cite{47761}:
{
\quoting[leftmargin=0.5cm,rightmargin=0.5cm]
    Q: {\it Who played will on as the world turns?} \\(A: {\it Jesse Soffer}.)\\  
    Q: {\it How many episodes in season 2 breaking bad?} (A: {\it 13}.)\\
    Q: {\it If a person who always tells lies tells you that the destination is on the left, should you go left or right?}  (A: {\it Right.})  
\endquoting
}
Answering these questions requires knowledge and logical reasoning capabilities in a variety of topics. 
The answers to these questions are often unique and clear,
answering them does not necessarily require creativity.

On the other hand, {\it open-ended} questions are more vague and abstract, and do not have clearly correct answers;
even experts with specialized knowledge will need to give consideration before giving an answer.
Examples of open-ended questions are 
{
\quoting[leftmargin=0.5cm,rightmargin=0.5cm]
    Q: {\it What are the best practices for conducting load tests on applications that expose REST APIs?}  \\ 
    Q: {\it What is a feature you would not like in your product?}  \\ 
      Q: {\it How can we prevent participants from arriving late to weekly meetings?}
\endquoting
}
The answers to these questions can vary widely. For instance, for the question, "How can we prevent participants from arriving late to weekly meetings?", the following are three possible answers: 
{
\quoting[leftmargin=0.5cm,rightmargin=0.5cm]
A: {\it Setting reminders: Set alerts or reminders 15 or 30 minutes before the meeting.} \\ 
A: {\it Time verification: Ensure that the meeting time is convenient for everyone and adjust if necessary.} \\ 
A: {\it Agenda sharing: Share the agenda the day before the meeting, allowing participants to prepare.}
\endquoting
}
Open-ended questions are essential for generating new ideas, encouraging creativity, and facilitating deeper understanding, but require detailed and multi-faceted consideration for evaluation. 
The superiority or inferiority of an answer depends on which evaluation criteria are used, and an overall decision must be made by synthesizing evaluations based on different evaluation criteria, which can be a very complex decision-making task.

In this study, we consider the use of LLMs to evaluate responses to open-ended questions.
In order to have LLMs evaluate answers from multiple perspectives, we focus on the Analytic Hierarchy Process (AHP)~\cite{SAATY1987161,saaty2004decision}, which is known as a mathematical decision-making tool in operations research, and execute the decision-making procedure using LLMs.
AHP evaluates candidates according to multiple evaluation criteria and makes a final decision by weighting them according to the relative importance of the evaluation criteria.
By having LLMs perform the AHP process, we expect to enhance the ability of LLMs to evaluate multiple aspects objectively.
Specifically, our proposed method could be divided into two phases: the criteria generation phase and the evaluation phase. We first have an LLM generate multiple evaluation criteria for answers to a question and determine the relative importance of these criteria through pairwise comparisons in the criteria generation phase.
We then perform pairwise comparisons of candidate answers for each criterion in the evaluation phase.
Finally, these results are integrated to arrive at a final decision. 



Our results demonstrate that multiple criteria can effectively evaluate answers to open-ended questions. We conducted experiments on 4 datasets on ChatGPT-3.5-turbo and GPT-4. Compared to baseline methods, our approach performs better on quantitative indicators,  concordance index, and soft concordance index. The contributions of this study are summarized as follows:
\begin{itemize}
\item 
We propose a new method that leverages LLMs to evaluate answers to open-ended questions. We conducted experiments using two different LLMs across four datasets.
\item Our findings indicate that our proposed method outperforms four baseline models. Notably, we observed that pairwise comparison plays a crucial role in assessing answers to open-ended questions.
\item We explored the impact of the number of evaluation criteria on the results. Our results demonstrate that using multiple criteria yields better performance.
\end{itemize}

%% file: related.tex
\section{Related Works}

\citet{reja2003open} examined human responses to the same question presented in two formats: open-ended and close-ended questions. The findings revealed that open-ended questions elicited a more diverse range of answers from participants. This suggests that open-ended questions play a more active role in exploring new ideas, requiring not just background knowledge and logical reasoning, but also a higher degree of intelligence and creativity.

\citet{belay2022ahp} proposed using the AHP to employ multicriteria analysis on the success factors of the Ethiopian construction industry, aiming to enhance decision-making capabilities in the construction sector.

\citet{svoboda2024enhancing} utilized the AHP and GPT-4 to generate answers for open-ended questions as an automated decision-making process. However, the proposed method lacked validation and quantitative evaluation metrics. Given the inherent uncertainty of open-ended questions, quantitative evaluation poses significant challenges. Our research focuses on developing and implementing quantitative evaluation metrics for our proposed methodology and evaluating the effectiveness of the combined use of AHP and GPT-4 in answering open-ended questions.

\citet{del2023automatic} summarized deep learning approaches for the automated scoring of answers of students to open-ended questions. Those methods utilize deep learning to extract representations of answers and employ supervised learning with ground truth scores. However, such approaches have two main drawbacks. First, they require training or at least fine-tuning the network for different questions. Second, they rely on the availability of ground truth labels.

Similarly, \citet{uto2020automated} proposed the use of LSTM networks for scoring answers of students to open-ended questions. Such supervised learning methodology could properly assess the ability of students but might not facilitate the discovery of new knowledge or identify better answers to open-ended questions without prior knowledge. The presence of ground truth labels implies that the superior answers are predetermined, limiting the potential for innovation in question answering. Our proposed method can be directly applied to LLMs without the need for training or ground truth labels. By quantitatively assessing the responses to open-ended questions, our approach allows for the discovery of new insights.

%% file: proposed.tex
\section{Methodology}

\subsection{Preliminaries: Analytic Hierarchy Process}



AHP is a decision-making technique through evaluating possible candidates under multiple evaluation criteria~\cite{SAATY1987161,saaty2004decision} and is extensively applied across various fields to tackle complex decision-making problems~\cite{liu2020review,bruno2012ahp,podvezko2009application,sari2017decision}. 

For example, when choosing a restaurant for dinner, we might consider several criteria such as the quality of the food, service, and price. To prioritize these criteria, we use the pairwise comparison method, which involves comparing them against each other. We might ask, "Which is more important: the quality of the food or the service?" to determine the relative importance of each criterion. Then, using the same pairwise comparison approach, we assess the restaurants under each criterion by asking questions such as "Does Restaurant A have better food than Restaurant B?". 

\begin{figure}[!t]
\centering
\includegraphics[width=0.99\linewidth]{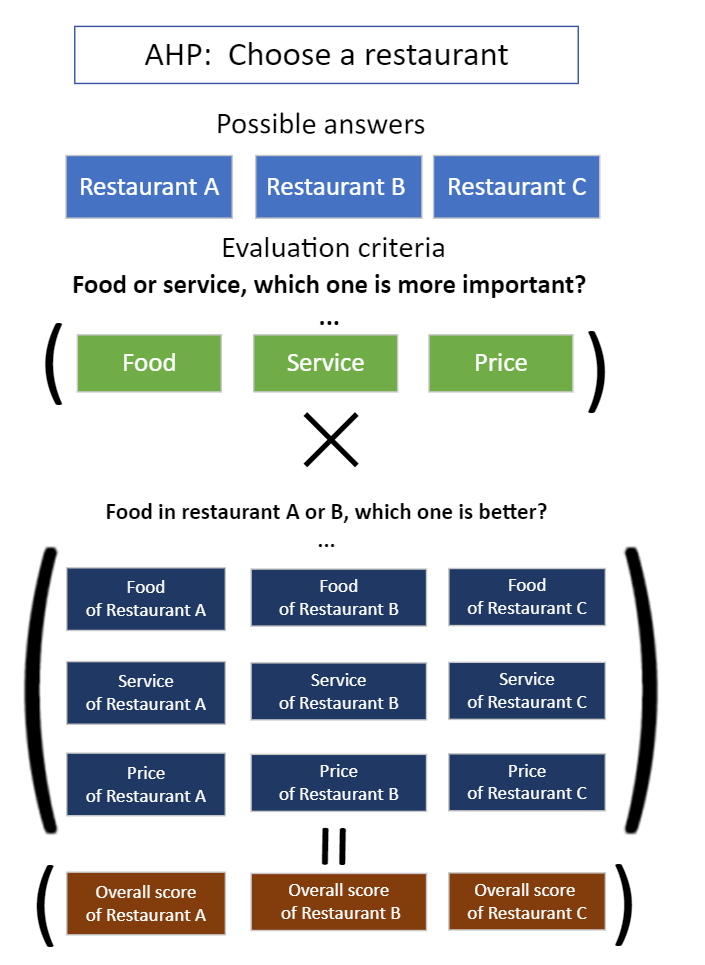}

\caption{
\textbf{An example of using AHP to choose a restaurant.} Three criteria: food, service, and price are used to decide which restaurant to choose. Each criterion has a weight representing its importance, and each restaurant has a score under each criterion, all obtained through pairwise comparison. The final overall scores can be obtained through matrix multiplication. 
}
\label{ahp}
\end{figure}

The overall scores for the restaurants are calculated by performing a matrix multiplication of the scores for each criterion with the corresponding weights of these criteria. Figure~\ref{ahp} illustrates this process.

\subsection{AHP-Powered LLM Reasoning for Evaluating Open-Ended Responses}

\begin{figure}[!t]
\centering
\includegraphics[width=0.85\linewidth]{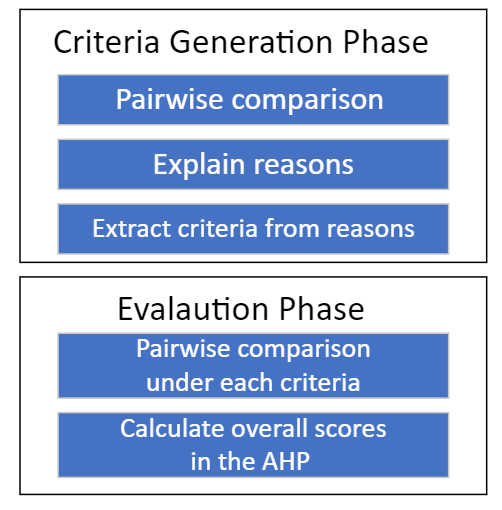}

\caption{
\textbf{Our proposed AHP-Powered LLM evaluation.}  
}
\label{ahppropose}
\end{figure}

Our proposed method which has LLMs automatically evaluate existing human open-ended responses, is divided into two distinct phases: the criteria generation phase and the evaluation phase. 
In the criteria generation phase, we generate multiple criteria to evaluate answers, tailored to capture the various dimensions of answer quality. In the subsequent evaluation phase, the answers are subjected to pairwise comparisons using the previously generated criteria. Figure~\ref{ahppropose} shows the steps of our proposed method.

\subsubsection*{Criteria Generation Phase}

We first randomly extract a small set, which contains $m$ responses, from each dataset, resulting in $_mP_2$ ordered pairs. For each pair, we ask LLMs to explain why the first answer in the pair is better than the second and to summarize $2$ or $3$ reasons. These reasons are subsequently used as the criteria for later evaluations. After querying the LLMs about all $_mP_2$ pairs, we typically gather around $\frac{5}{2}(_mP_2)$ reasons. Many of these reasons are repetitive. We input all collected reasons back into the LLMs, which then rank and output the top $k$ reasons according to their importance and frequency of occurrence. We use these top $k$ reasons as $k$ evaluation criteria.

\subsubsection*{Evaluation Phase}
We utilize the previously prepared $k$ evaluation criteria to conduct pairwise comparisons on all answers under each criterion. For a given answer pair $i$ and $j$, we provide five options for the LLMs to select from: "answer $i$ is better than answer $j$", "answer $i$ is slightly better than answer $j$", "almost the same", "answer $j$ is slightly better than answer $i$", "answer $j$ is better than answer $i$". Under each evaluation criterion, we can construct a $k \times n \times n $ tensor $\mathcal{A}$, which are the outcomes of pairwise comparisons, where $n$ is the number of responses. We have simplified the original nine choices of the AHP to only five options for the LLMs to choose from. The definition of tensor $\mathcal{A}$ is as follows: 
\begin{equation}
    \mathcal{A}_{kij} =\begin{cases}
  5 &\text{if }\ i \gg j  \text{ under criteria} \ k, \\
  3 &\text{if }\ i  > j  \text{ under criteria} \ k, \\
  1 &\text{if }\  i=j  \text{ under criteria} \ k, \\
  \frac{1}{3} &\text{if }\ j \gg i  \text{ under criteria} \ k, \\
  \frac{1}{5}  &\text{if }\ j > i  \text{ under criteria} \ k ,
 
\end{cases}
\end{equation}
where $\gg$ refers to `better than', $>$ refers to `slightly better than' and $i=j$ refers to `almost the same', respectively. Numbers `$1$', `$3$', and `$5$' are an example of specified numbers in the AHP, which can be different numbers in the other cases.

Next, we calculate the weights for each criterion by first constructing a preference $k \times k $ matrix $\mathcal{W}$:
\begin{equation}
\mathcal{W}_{pq} =\begin{cases}

  3 &\text{if }\ p<q, \\
  1 &\text{if }\ p=q, \\
  \frac{1}{3} &\text{if }\ p>q. \\
\end{cases}
\end{equation}
Since we have already asked the LLMs to rank the $k$ criteria by importance during their generation, we assume that earlier criteria are more important than later ones. Differing from typical AHP, we did not perform pairwise comparisons between criteria as LLMs tend to produce output "almost the same" in the importance comparisons of criteria.

Next, we calculate the $k$-dimensional vector $\bm{w}=\sigma^{\downarrow}_{1}(\mathcal{W})$, where $\sigma^{\downarrow}_{1}(.) $ returns the eigenvector corresponding to the largest eigenvalue of the input matrix.
We use the normalized vector $\bm{\widetilde{w}}$ as the weights for the criteria, where
\begin{equation}
   \widetilde{\bm{w}}_i =  \frac{\bm{w}_i}{\sum_{j=1}^n \bm{w}_j}.
\end{equation}
The scores for each answer under each criteria $k \times n $ matrix $\mathcal{S} $ can be calculated as 

\begin{equation}
   \mathcal{S}_k = \sigma^{\downarrow}_{1}(\mathcal{A}_k),
\end{equation}
and normalized scores under each criteria $ \mathcal{\widetilde{S}} $ can be calculated as 
\begin{equation}
    \mathcal{\widetilde{S}}_{ki}  = \frac{\mathcal{S}_{ki}}{\sum_i \mathcal{S}_{ki}}.
\end{equation}
Finally, we combine all the criteria to calculate the final score for each answer,

\begin{equation}
    \bm{s} =  \mathcal{S}^T \widetilde{\bm{w}}_i.
\end{equation}

%% file: results.tex
\begin{table*}[!]

    \centering
    \resizebox{1\linewidth}{!}{
    \begin{tabular}{l|l|c|l}
    \toprule
       Dataset & Question & $\#$Responses & Ground Truth \\
        \midrule
         Part-time job  & It is important for college students to have a part-time job. & 80 & CEFR level\\ 

         Smoking  & Smoking should be completely banned at all the restaurants in the country. & 80 & CEFR level\\ 
          Meeting  & How can we reduce the number of latecomers for team meetings? & 80 & Human evaluation \\ 
           Cheat  & How can we effectively prevent students from cheating in exams? & 80 & Human evaluation \\ 
        \bottomrule
    \end{tabular}
    }
    
      \caption{
      \textbf{Four datasets used in the experiments.} The "Question" column describes the open-ended problems that need to be addressed. We assume that a good automated evaluation method should more closely resemble the human evaluations in the "Ground Truth", thereby quantifying the effectiveness of our proposed method. }
    
\label{dataset}
\end{table*}

\begin{table}[!t]
\small
    \centering
   
    \begin{tabular}{l|c}
    \toprule
       Method & $\#$Requests with LLMs \\
        \midrule
         AHP (Ours)  &  $  ( kn(n-1) ) /2 $\\ 

         Pairwise  &  $ (n(n-1))/2 $\\ 
          Scoring  &  $ n $\\ 
           Few-shot  &  $ n $\\ 
            Level  &  $ n $\\ 
        \bottomrule
    \end{tabular}

      \caption{
      {
      \textbf{Number of Requests to LLMs of our proposed method and the baselines.} $k$ refers to the number of criteria and $n$ the size of the dataset. Our method requires more requests. However, since the dataset size is 80, which is relatively small, the time and economic costs are acceptable for LLMs. In most real-world scenarios, 80 answers for one open-ended question are relatively sufficient.
      }}
    
\label{query}
\end{table}


\begin{table*}[!t]
\small
\centering

\begin{tabular}{l|ccccccccc}
\toprule
Method & \textbf{AHP (Ours)}  &  Pairwise Comparison   &  Scoring  &  Few-shot &  level   \\ \midrule

 ~ & \multicolumn{5}{c}{ChatGPT-3.5-turbo}     \\ \midrule

Part-time job  & 69.1 & 64.7   &  41.3  &  25.4  &   30.9  \\ 
Smoking   & 78.6 & 75.3   &  51.7  & 27.8  &   53.7  \\  
Meeting  & 61.3 &  62.9  &  36.8  & 43.6  &   N/A  \\ 
Cheat   & 63.6 & 56.2   &  45.3 & 43.9  &   N/A  \\  

\midrule

 ~ & \multicolumn{5}{c}{GPT-4}     \\ \midrule

Part-time job  & 72.6 & 66.9   &  44.1  & 24.2  &   16.8  \\ 
Smoking   & 79.1 & 76.0   & 57.6   & 10.4   &   42.2  \\  
Meeting  & 65.4 & 65.3 & 50.6    &  25.1  &    N/A   \\ 
Cheat   & 72.6 & 61.6  &  49.6    &  23.5  &    N/A  \\  

\bottomrule

\end{tabular}

\caption{
{
\textbf{Results of concordance index. } Our proposed AHP method outperforms all baselines except for the Meeting using ChatGPT-3.5-turbo. Our multi-criteria approach effectively enhances the performance of pairwise comparisons. }}

\label{GPT3main1}
\end{table*}

\begin{table*}[!t]
\small
\centering

\begin{tabular}{l|ccccccccc}
\toprule
Method & \textbf{AHP (Ours)}    &  Pairwise Comparison   &  Scoring  &  Few-shot &  level   \\ \midrule

 ~ & \multicolumn{5}{c}{ChatGPT-3.5-turbo}     \\ \midrule

Part-time job  & 75.3 & 69.1   &  41.4 & 23.9  &   32.6  \\ 
Smoking   & 87.8 & 83.8   &  62.3  & 32.1  &   62.7  \\  
Meeting  & 66.4 & 68.2   &  39.9  & 50.1  &    N/A   \\ 
Cheat   & 70.6 & 59.6   &  52.9  & 51.4  &    N/A   \\

\midrule

 ~ & \multicolumn{5}{c}{GPT-4}     \\ \midrule

Part-time job  & 78.7 & 70.9   &  49.8  & 26.1  &   19.9  \\ 
Smoking   & 86.8 & 86.9   & 65.8   &  11.4  &   51.2  \\  
Meeting  & 72.7 & 70.4  & 56.7    &  31.1  &   N/A  \\ 
Cheat   & 84.0 & 67.7  &  56.4     &  27.5  &    N/A   \\ 
\bottomrule

\end{tabular}

\caption{
{
\textbf{Results of soft concordance index.} Our proposed method performs better than other methods under easier metrics.  }
}
\label{GPT4main}
\end{table*}
















\section{Experiments}

We conducted experiments to answer the following three research questions:
\begin{itemize}
\item [RQ1] Could our proposed method effectively evaluate open-ended responses?
\item [RQ2] What is the impact of multiple criteria on the results?
\item [RQ3]What are the findings in experiments with different LLMs?
\end{itemize}

\subsection{Experiment Settings}

We utilized four datasets, Part-time job and Smoking are from The International Corpus Network of Asian Learners of English (ICNALE)~\cite{ishikawa2018icnale}, while Meeting and Cheat are focused on proposing ideas for practical issues~\cite{baba2020crowdea,crowdopinion,pramoe}. We randomly selected $80$ human responses to open-ended questions from each dataset. 

In the Part-time job and Smoking datasets, there are English essays written by students at four different CEFR levels\footnote{\url{https://www.coe.int/en/web/common-european-framework-reference-languages/table-1-cefr-3.3-common-reference-levels-global-scale}}, with 20 individuals per level. We use the English CEFR level as the ground truth, where individuals with higher CEFR levels should receive higher scores.

In the Meeting and Cheat datasets, all 80 ideas were annotated by humans on the crowdsourcing platform, forming a human-evaluated ranking. Ideas rated higher by humans should receive higher scores.

We evaluate the effectiveness of each evaluation method by comparing their deviations from the ground truth. The more effective an evaluation method is, the closer it should align with the ground truth. The dataset settings are summarized in Table~\ref{dataset}.

We empirically extract $10$ responses, form $_{10}P_2=90$ ordered pairs, have LLMs generating around $225$ reasons, and have LLMs summarize $10$ criteria for each dataset. For example, for Part-time jobs on ChatGPT-3.5-turbo, 10 criteria are ranked by importance as follows: "Clarity and Coherence", "Depth of Analysis", "Use of Evidence and Examples", "Grammar and Language Proficiency", "Logical Argumentation", "Structure and Coherence", "Argument Development", "Critical Thinking and Analysis", "Supporting Details and Examples" and "Use of Personal Experience".

\subsection{Baselines}

To quantitatively compare our proposed method, our experiments include four baselines, summarized as follows:

\begin{itemize}
\item Pairwise Comparison: Unlike our proposed AHP-powered multicriteria evaluation, this baseline involves having LLMs perform direct pairwise comparisons, deciding which answer is better without any criteria. 

\item Scoring: We have LLMs score answers on a scale from 0 to 100.

\item Few-shot In-context Learning: We ask LLMs to assign a level to each answer based on two given examples. In the Part-time job and Smoking dataset, we select two answers from each of the four levels as examples. In the Meeting and Cheat dataset, we extract the two best answers, two answers from the top 33\%, two from the top 66\%, and the two worst answers to serve as examples for four levels.

\item CEFR Level: We instruct LLMs to evaluate answers based on CEFR definitions, such as for CEFR B2 writing level, which is defined as "Can write clear, detailed texts on different subjects. Can use information and arguments from other sources in their writing." This baseline is used only in the Part-time job and Smoking dataset.
\end{itemize}

Our proposed method, compared to other baselines, requires more LLM queries, which is summarized in Table~\ref{query}.
We performed $108079/2 = 31600$ comparisons for each dataset. Each comparison contains roughly $300$ tokens, therefore the total input for each dataset is about 9.4 million tokens. The price for ChatGPT-3.5-turbo API was USD $3$ per million input tokens and USD $5$ per million input tokens for GPT-4o. We believe that the cost is relatively affordable and expect that future developments for saving LLM costs will make the use of the proposed method easier.

We use the concordance index to measure the discrepancy between the scores given by the evaluation method for each answer and the scores provided in the ground truth. The concordance index is defined as follows,

\begin{equation}
  \resizebox{0.45\textwidth}{!}{$
    \text{CI}(f,g) = \frac{\sum_i \sum_j{ \mathbb{I}(f(\bm{x}_i) >f(\bm{x}_j))\mathbb{I}(g(\bm{x}_i) >g(\bm{x}_j)) }}{\sum_i \sum_j{ \mathbb{I}(g(\bm{x}_i) >g(\bm{x}_j)) } }   $} ,
\end{equation}
where $\bm{x}$ is the dataset, $f$ is the evaluation method, $g$ is the ground truth and $\mathbb{I}$ is the indicator function.

When using the concordance index, some answers with minor differences may also be included in the calculation, therefore, we introduce soft as another metric that only takes answers with large differences into consideration and is defined as follows,
\begin{equation}
  \resizebox{0.45\textwidth}{!}{$
    \text{sCI}(f,g) = \frac{\sum_i \sum_j{ \mathbb{I}(f(\bm{x}_i) >f(\bm{x}_j))\mathbb{I}(g(\bm{x}_i) \gg g(\bm{x}_j)) }}{\sum_i \sum_j{ \mathbb{I}(g(\bm{x}_i) \gg g(\bm{x}_j)) } }   $} ,
\end{equation}
where $\gg$ represents a significant difference between two answers. In the Part-time job and Smoking dataset, It indicates that the level difference is greater than or equal to $2$. In the Meeting and Cheat dataset, it indicates that rankings differ by at least 20 positions within the dataset consisting of a total of 80 responses.

\begin{figure*}[!t]
\centering

\subfloat[Part-time job / 3.5]{\includegraphics[width=0.23\linewidth]{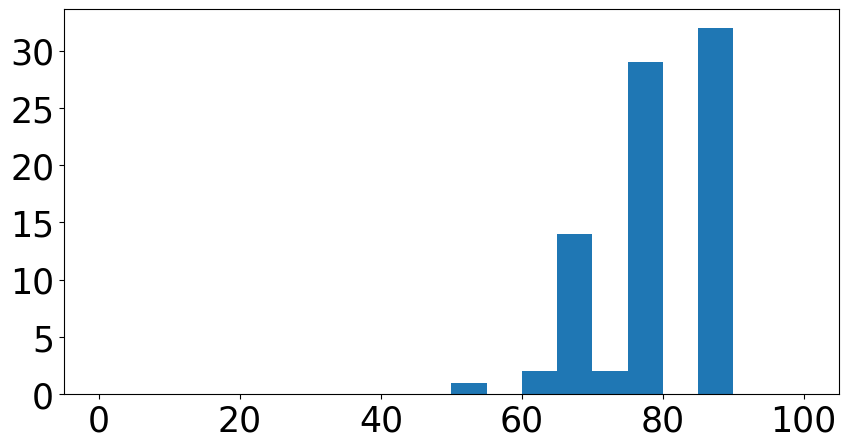} } 
\subfloat[Smoking / 3.5]{\includegraphics[width=0.23\linewidth]{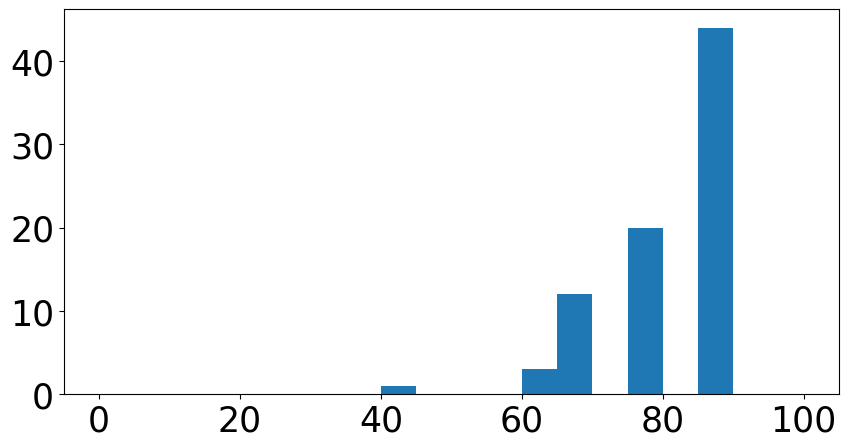} }  
\subfloat[Meeting / 3.5]{\includegraphics[width=0.23\linewidth]{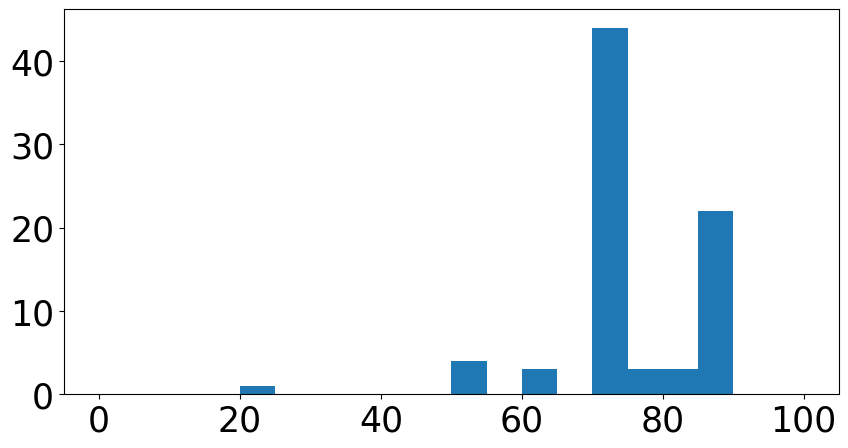} }
\subfloat[Cheat / 3.5]{\includegraphics[width=0.23\linewidth]{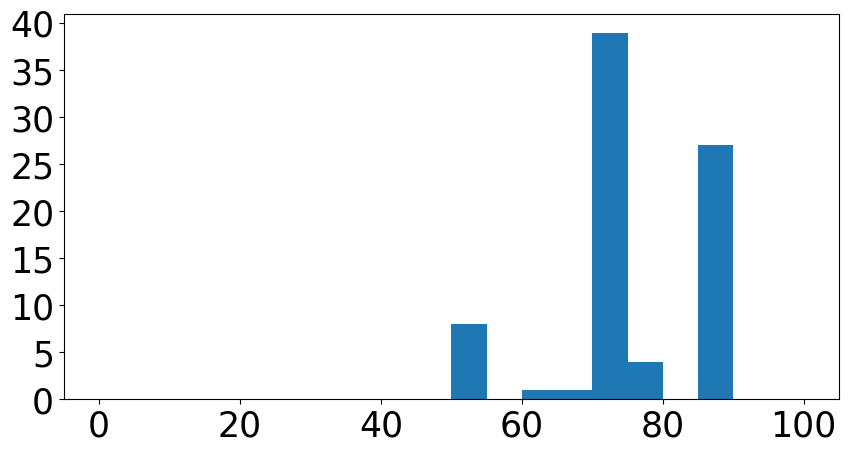} } \\

\subfloat[Part-time job / 4]{\includegraphics[width=0.23\linewidth]{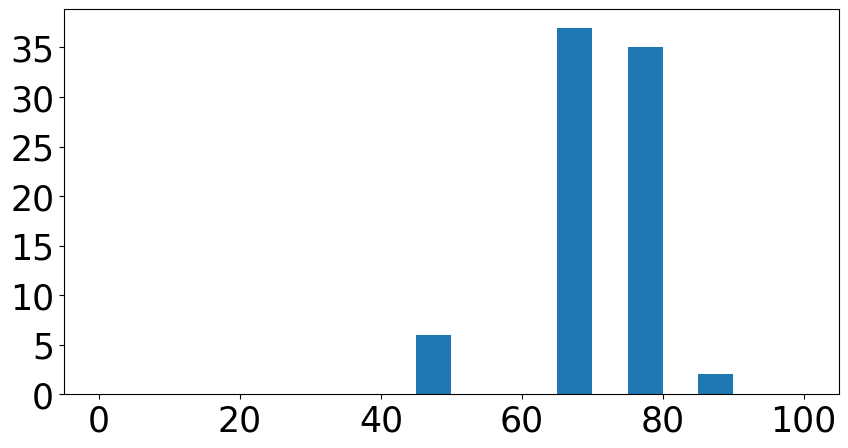} } 
\subfloat[Smoking / 4]{\includegraphics[width=0.23\linewidth]{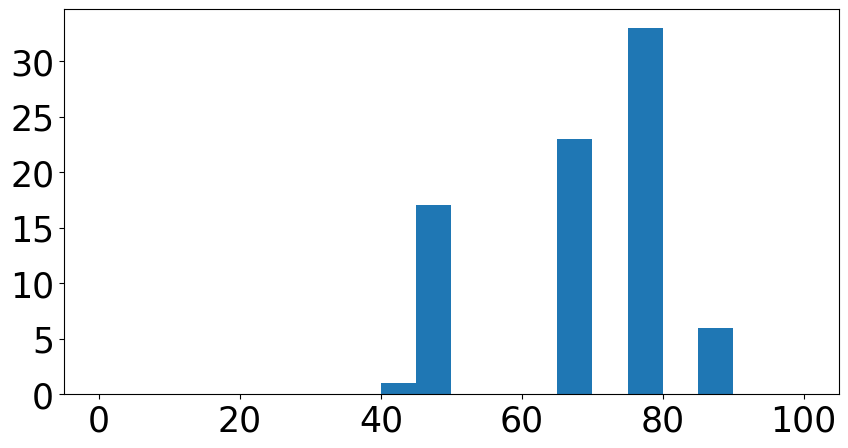} }  
\subfloat[Meeting / 4]{\includegraphics[width=0.23\linewidth]{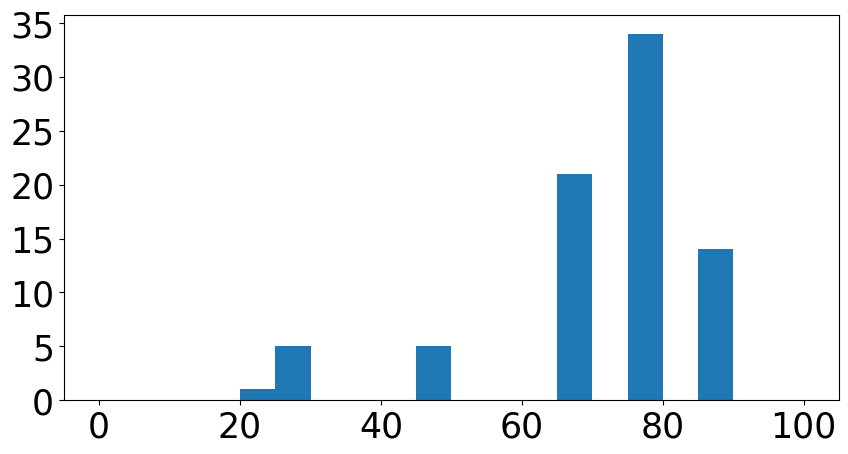} }
\subfloat[Cheat / 4]{\includegraphics[width=0.23\linewidth]{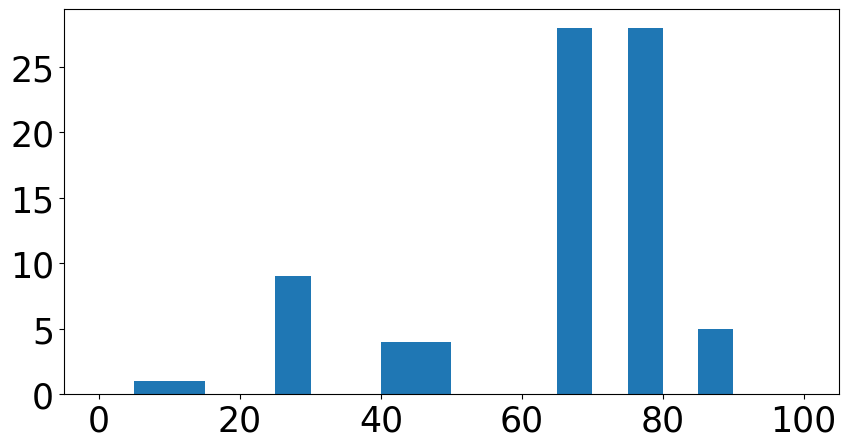} }

\caption{
\textbf{Histograms of Scoring evaluation.} The left of the `/' in the subtitle corresponds to the dataset, and the "3.5" on the right refers to ChatGPT3.5-turbo, while "4" refers to GPT-4. We will use the same notation in other figures. The horizontal and vertical axes represent scores and number of responses, respectively. It is shown that LLMs tend to assign mid to high scores to answers, which leads to a lack of differentiation and worsens the results. GPT-4 performs slightly better than ChatGPT-3.5 but still falls short of being satisfactory. 
}
\label{Scoring}
\end{figure*}

\begin{figure*}[!t]
\centering

\subfloat[Part-time job / 3.5]{\includegraphics[width=0.23\linewidth]{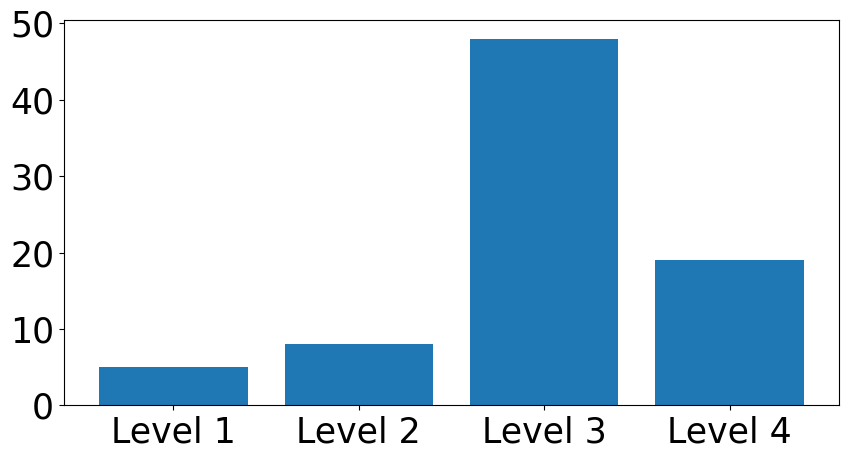} } 
\subfloat[Smoking / 3.5]{\includegraphics[width=0.23\linewidth]{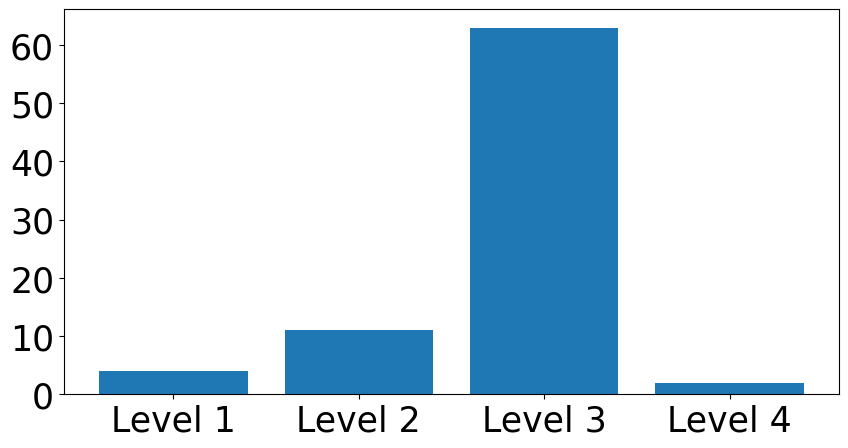} }  
\subfloat[Meeting / 3.5]{\includegraphics[width=0.23\linewidth]{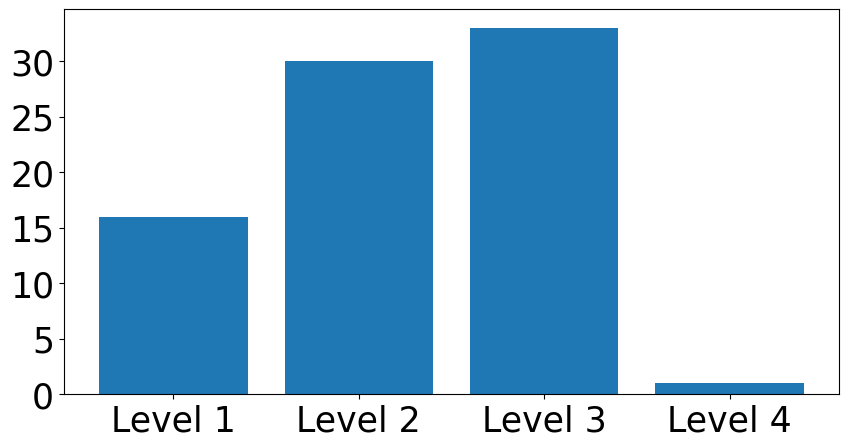} }
\subfloat[Cheat / 3.5]{\includegraphics[width=0.23\linewidth]{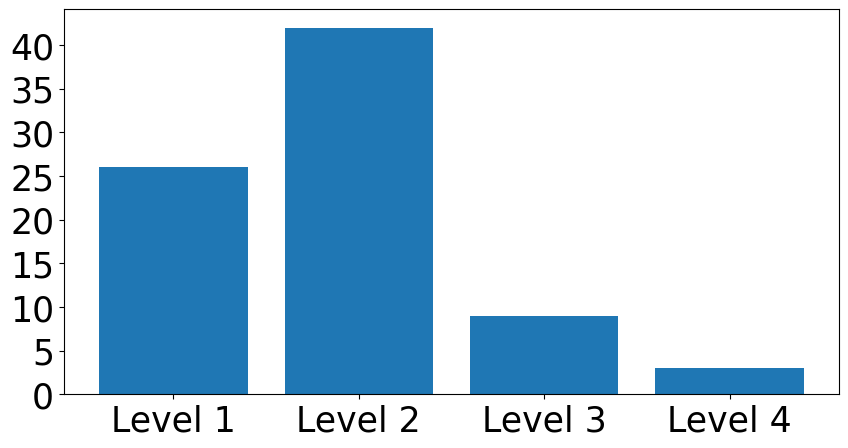} } \\

\subfloat[Part-time job / 4]{\includegraphics[width=0.23\linewidth]{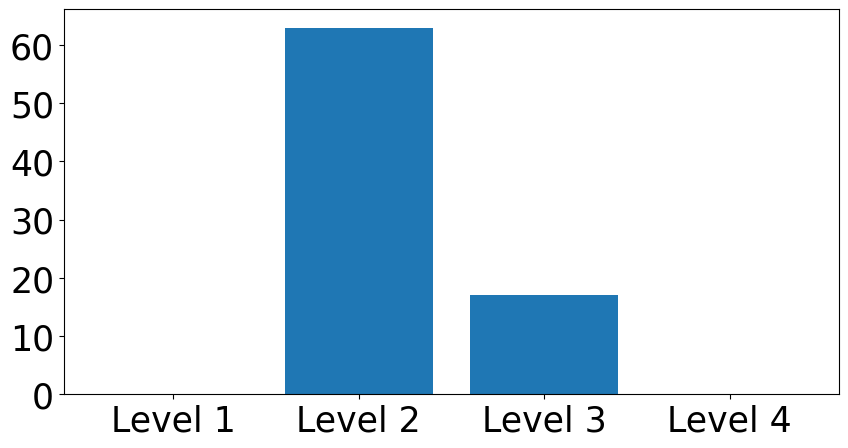} } 
\subfloat[Smoking / 4]{\includegraphics[width=0.23\linewidth]{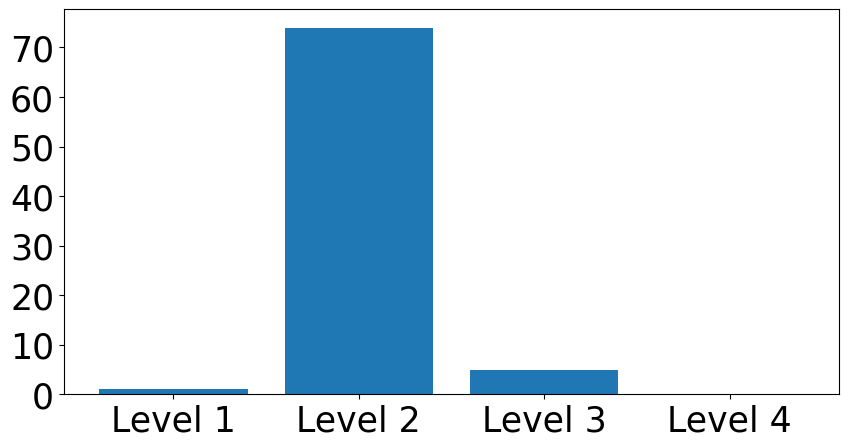} }  
\subfloat[Meeting / 4]{\includegraphics[width=0.23\linewidth]{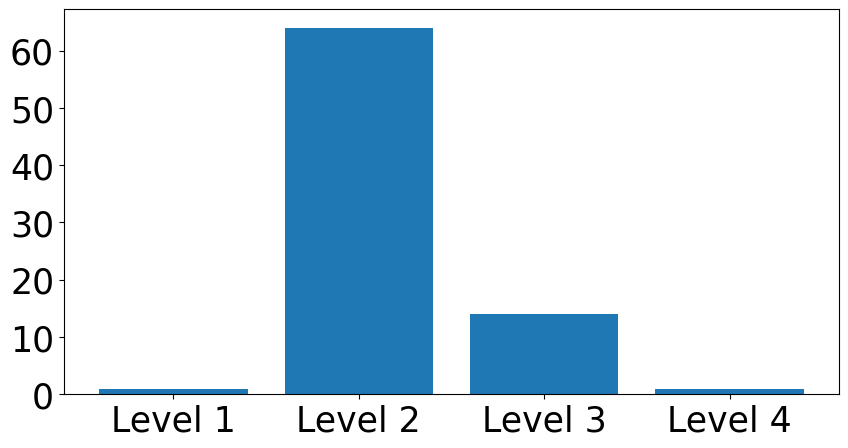} }
\subfloat[Cheat / 4]{\includegraphics[width=0.23\linewidth]{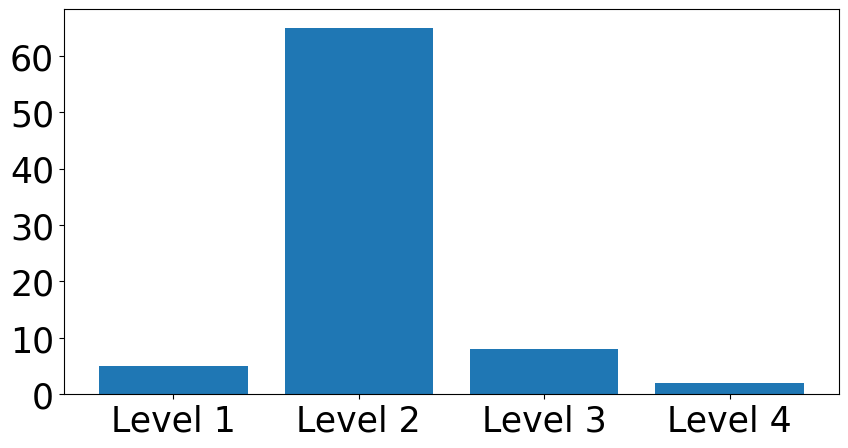} }

\caption{
\textbf{Histograms of Few-shot evaluation.} The horizontal and vertical axes represent the level and number of responses, respectively. It is shown that LLMs have almost no ability to learn from a small number of samples in complex open-ended questions. In most cases, LLMs tend to assign mid to high levels, while rarely assigning the highest or lowest levels. 
}
\label{Fewshot}
\end{figure*}

\begin{figure*}[!t]
\centering

\subfloat[Part-time job / 3.5]{\includegraphics[width=0.23\linewidth]{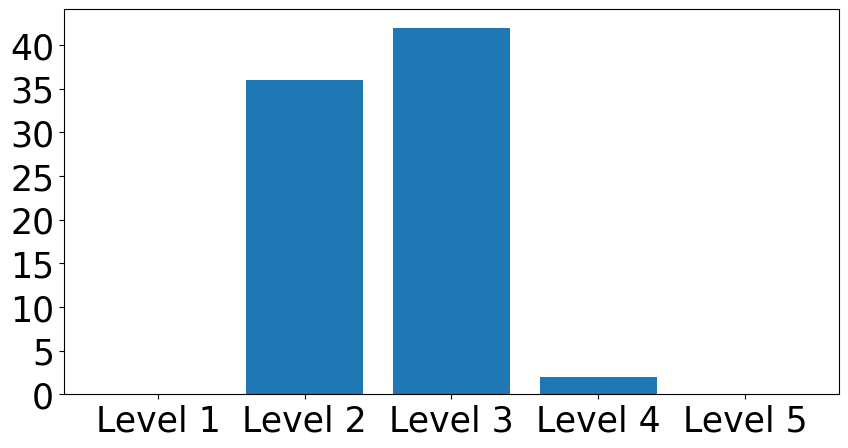} } 
\subfloat[Smoking / 3.5]{\includegraphics[width=0.23\linewidth]{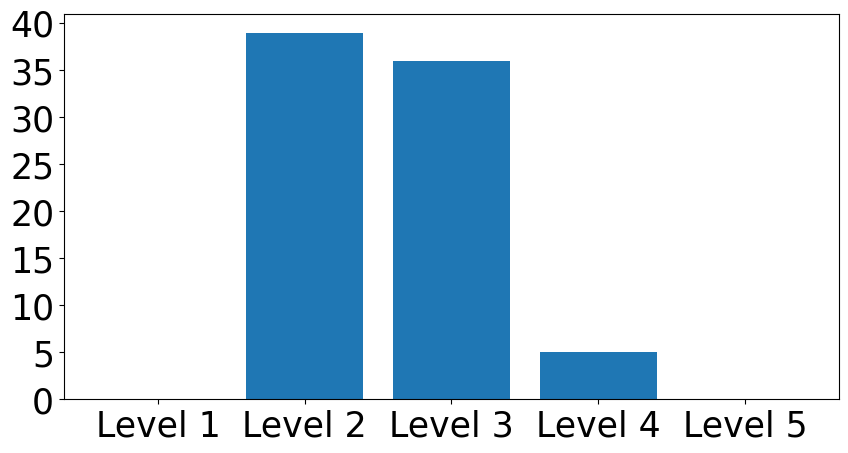} }
\subfloat[Part-time job / 4]{\includegraphics[width=0.23\linewidth]{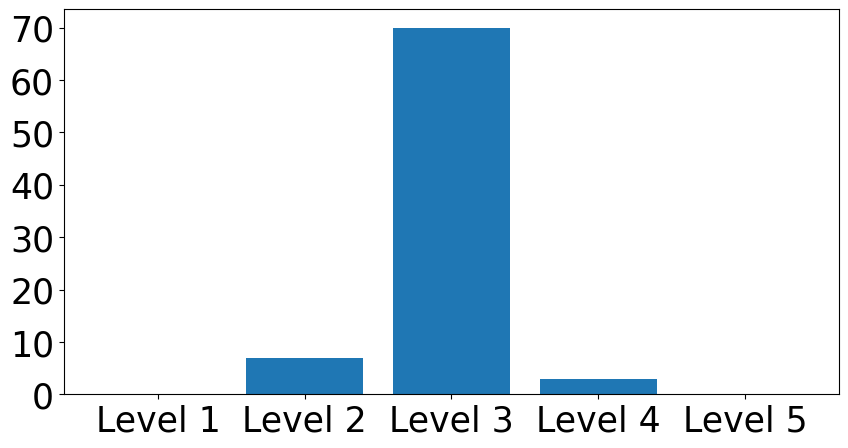} }
\subfloat[Smoking / 4]{\includegraphics[width=0.23\linewidth]{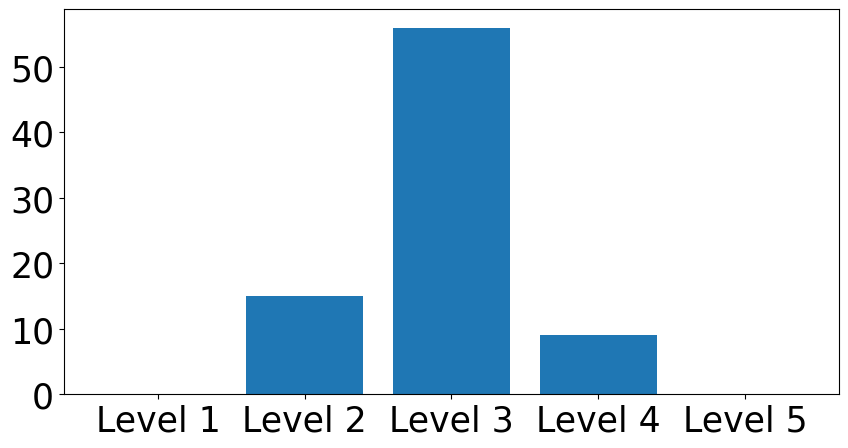} }

\caption{
\textbf{Histograms of CERF level evaluation.}  The horizontal and vertical axes represent the level and number of responses, respectively. It is shown that LLMs lack the ability to learn level definitions, tending to assign most articles a level of 2 or 3.
}
\label{CERF}
\end{figure*}

\subsection{RQ1: Could our proposed method effectively evaluate open-ended responses? -- Yes. }
Tables~\ref{GPT3main1} and \ref{GPT4main} show the results of the experiments. They indicate that even without multiple criteria, pairwise comparison significantly outperforms other baselines. This demonstrates that pairwise comparison is crucial for complex open-ended questions, as other baselines fail to elicit sufficiently effective information from LLMs. Figures~\ref{Scoring}, \ref{Fewshot}, and \ref{CERF} demonstrate that without pairwise comparisons, LLMs are unable to effectively and appropriately evaluate the quality of open-ended responses. Our proposed AHP-powered multicriteria evaluation outperforms all baselines except for the Meeting dataset with ChatGPT-3.5 while the performance of our proposed method is slightly below that of pairwise comparison.

\begin{figure*}[!t]
\centering

\subfloat[Part-time job / 3.5]{\includegraphics[width=0.23\linewidth]{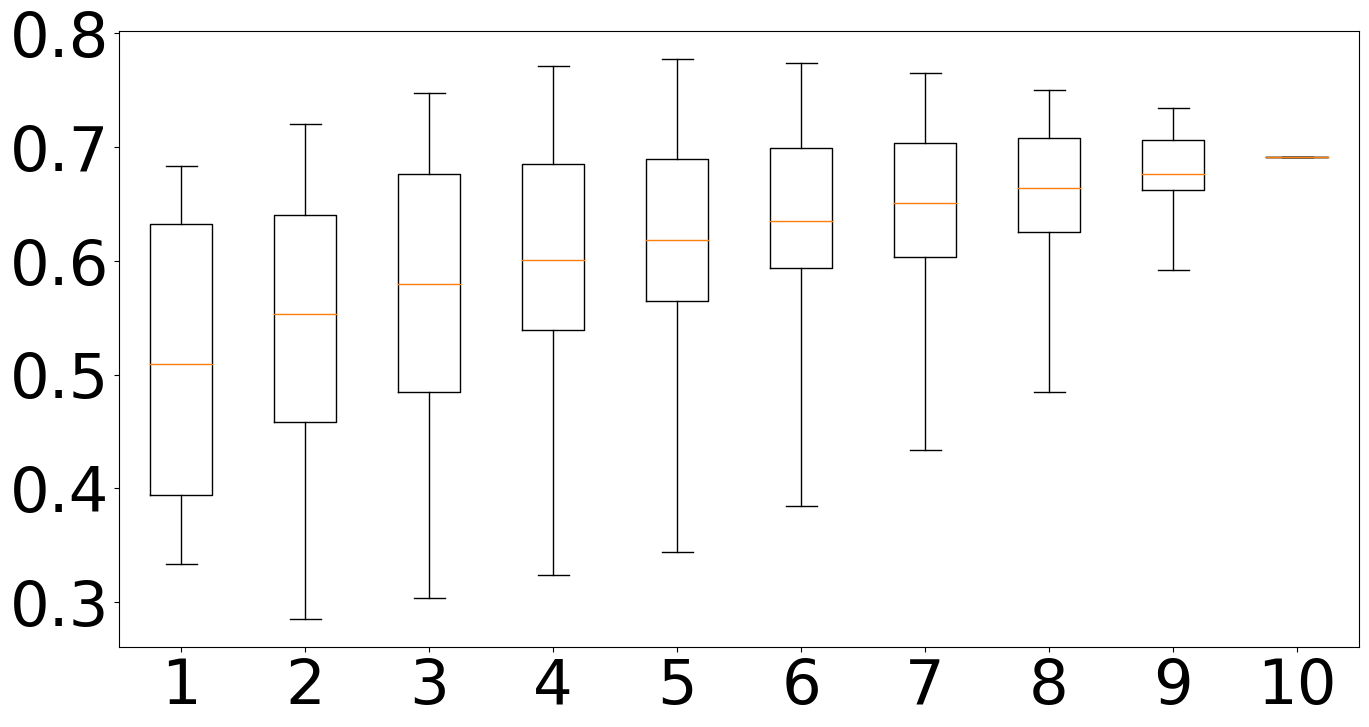} }
\subfloat[Smoking / 3.5]{\includegraphics[width=0.23\linewidth]{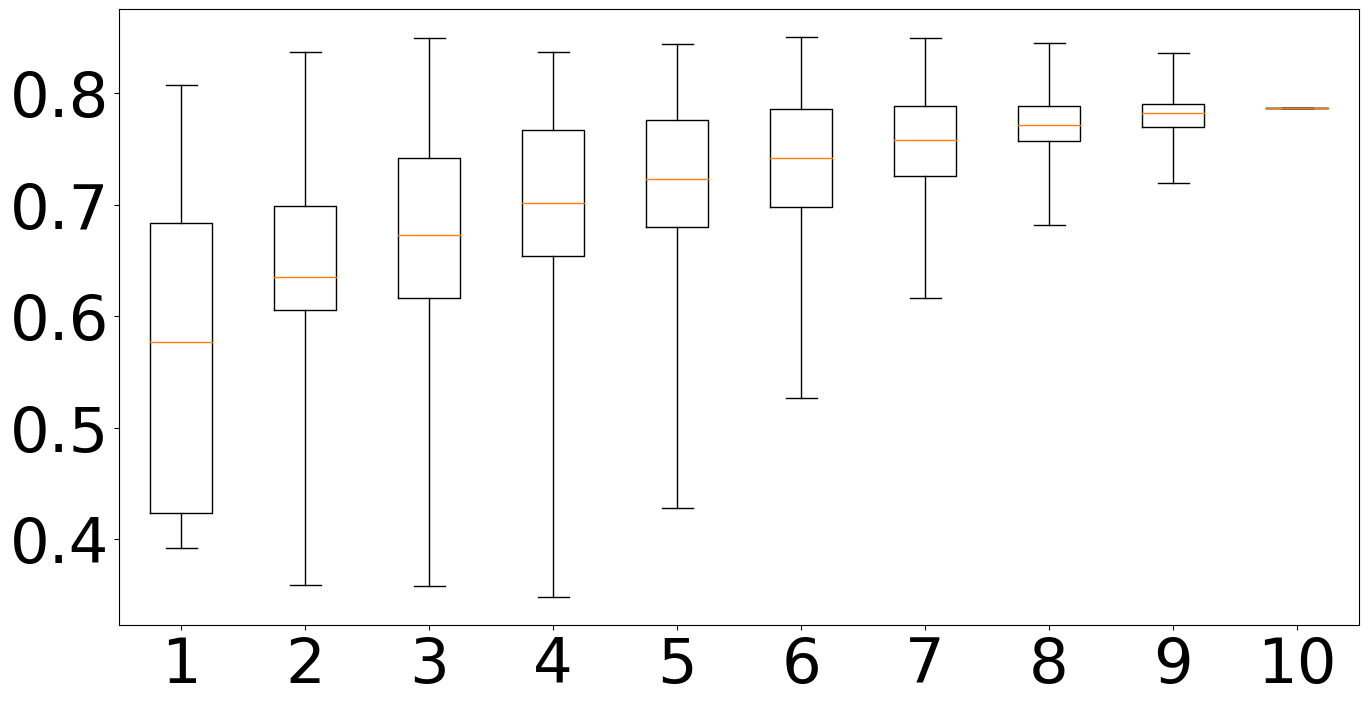} } 
\subfloat[Meeting / 3.5]{\includegraphics[width=0.23\linewidth]{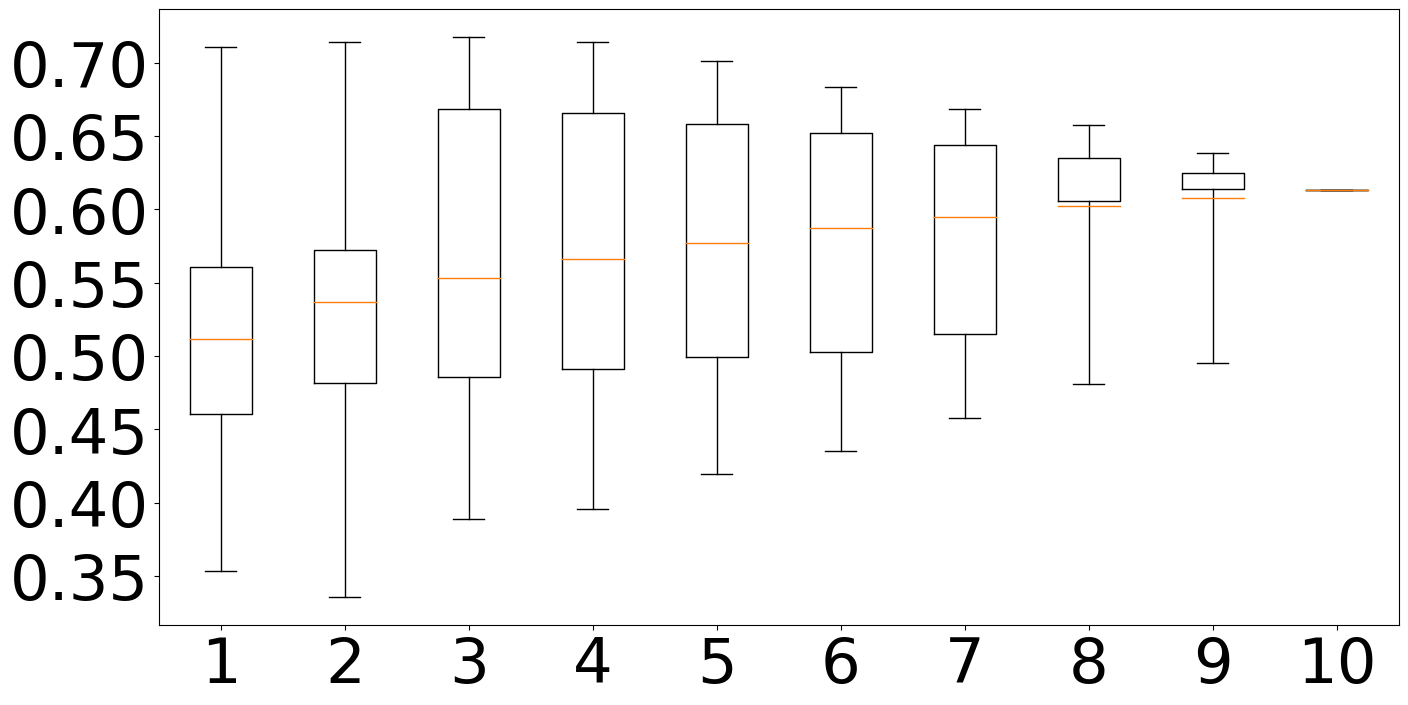} }
\subfloat[Cheat / 3.5]{\includegraphics[width=0.23\linewidth]{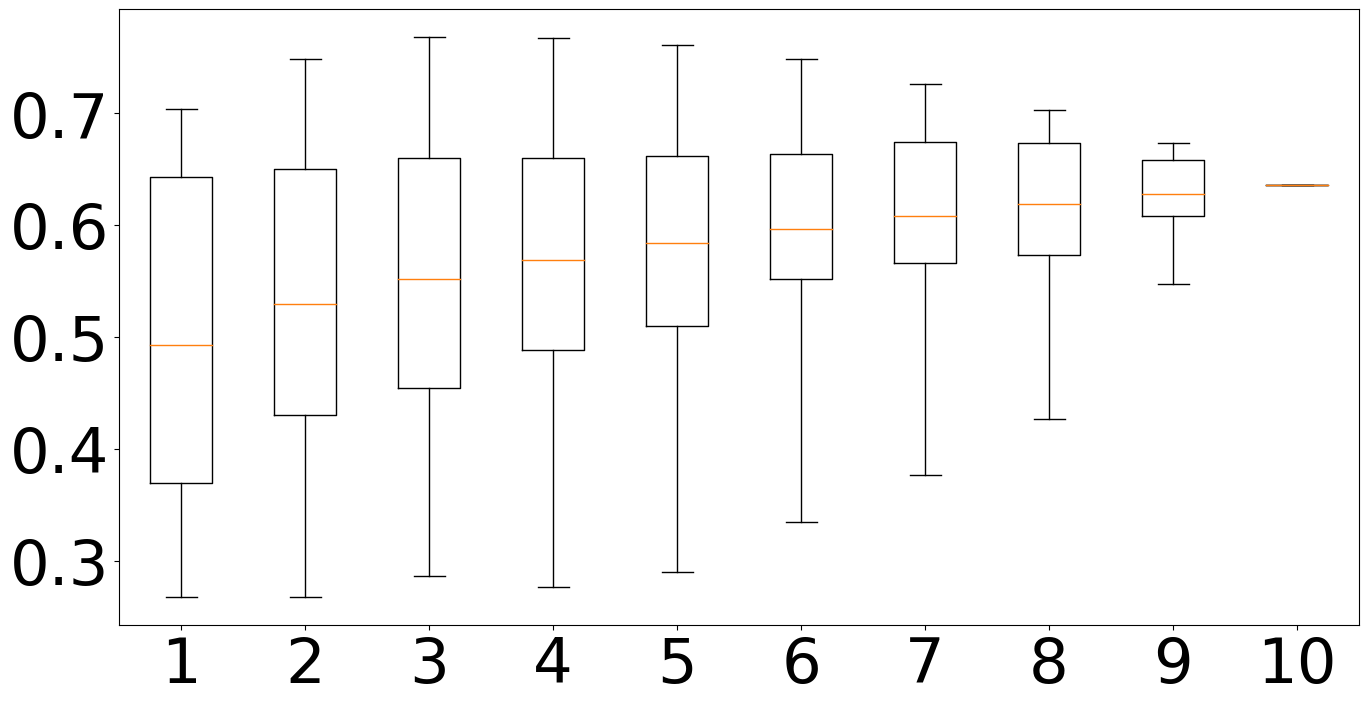} } 

\subfloat[Part-time job / 4]{\includegraphics[width=0.23\linewidth]{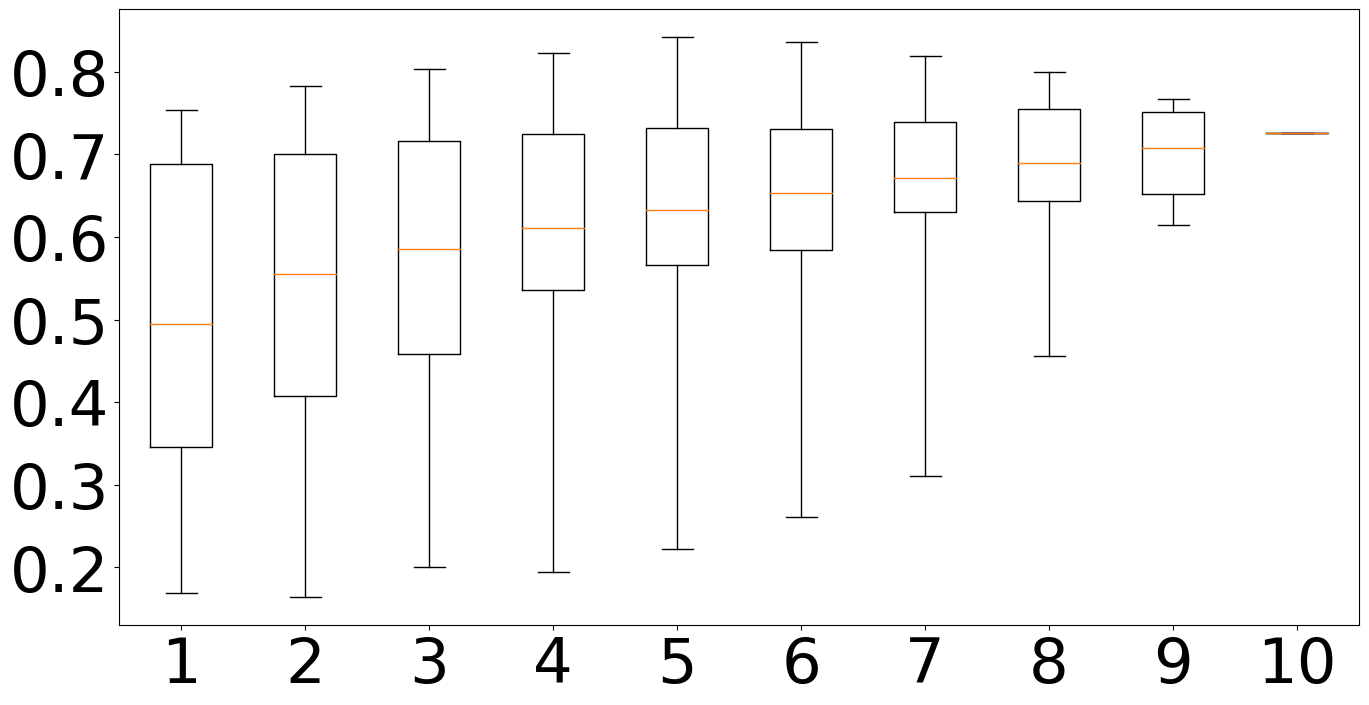} }
\subfloat[Smoking / 4]{\includegraphics[width=0.23\linewidth]{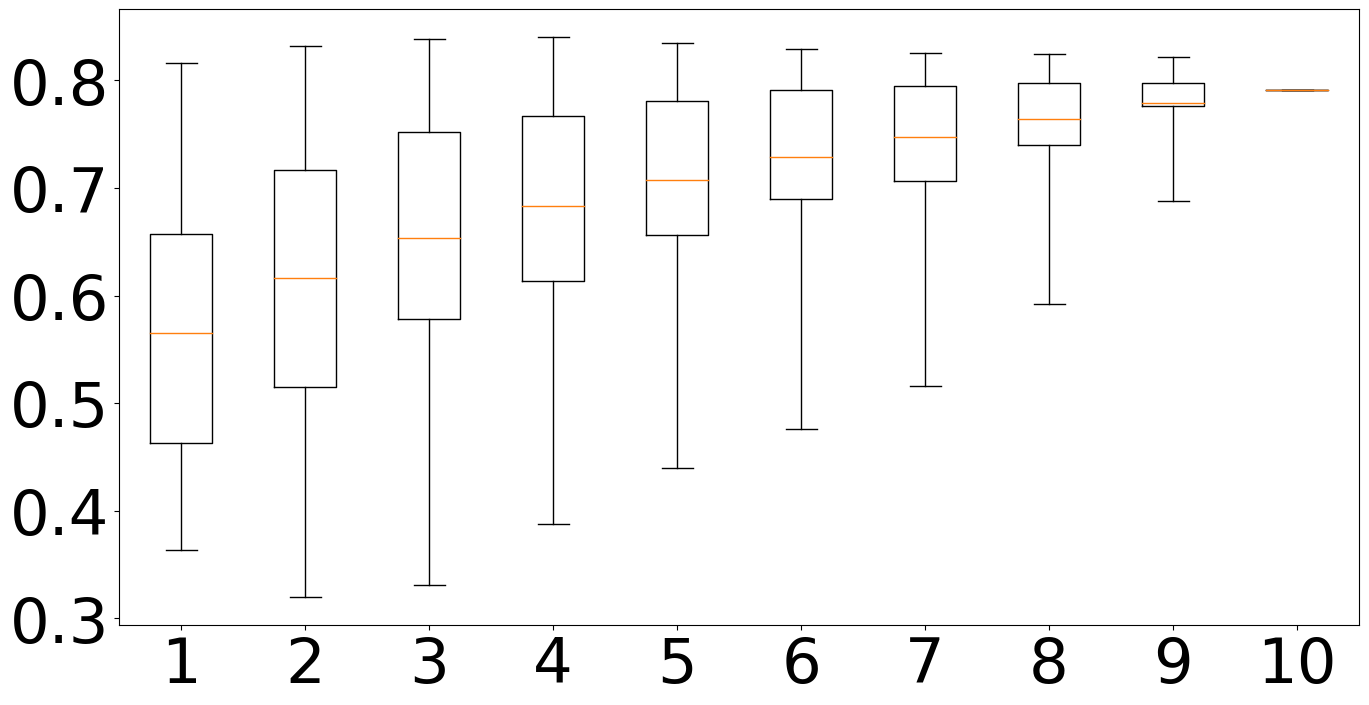} } 
\subfloat[Meeting / 4]{\includegraphics[width=0.23\linewidth]{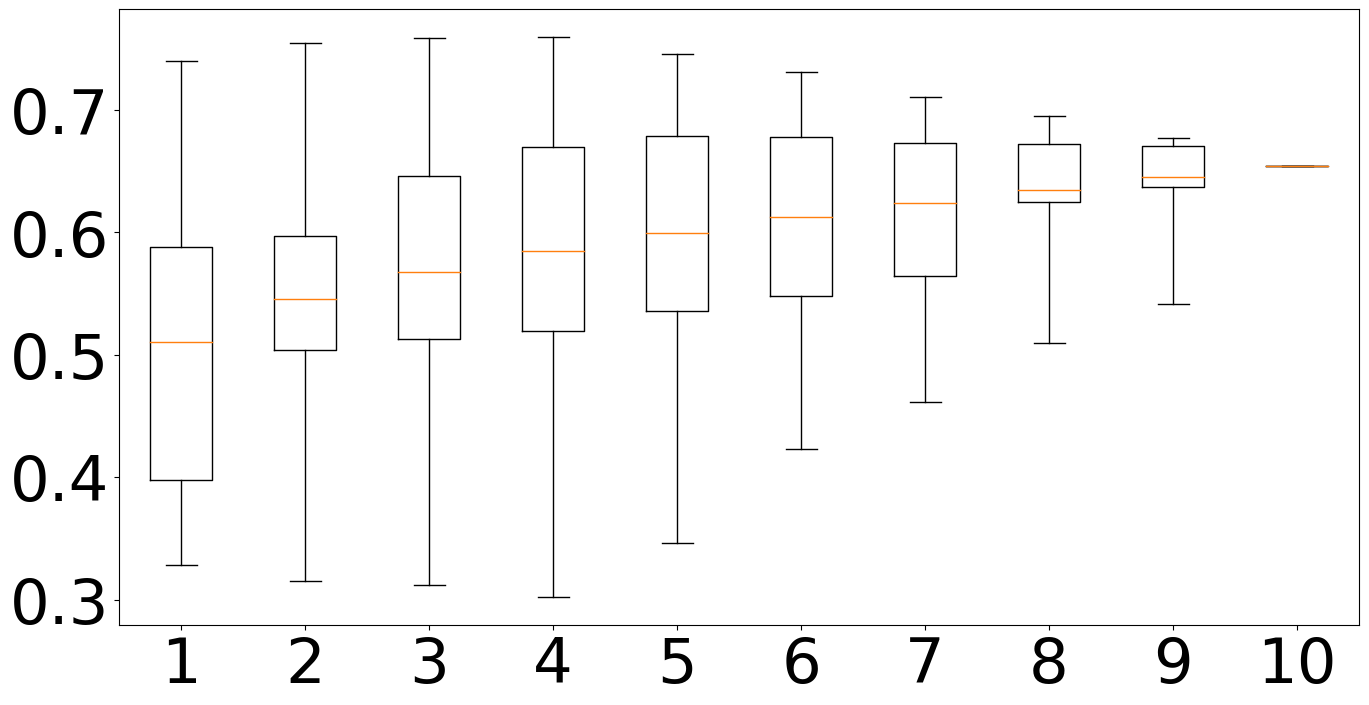} }
\subfloat[Cheat / 4]{\includegraphics[width=0.23\linewidth]{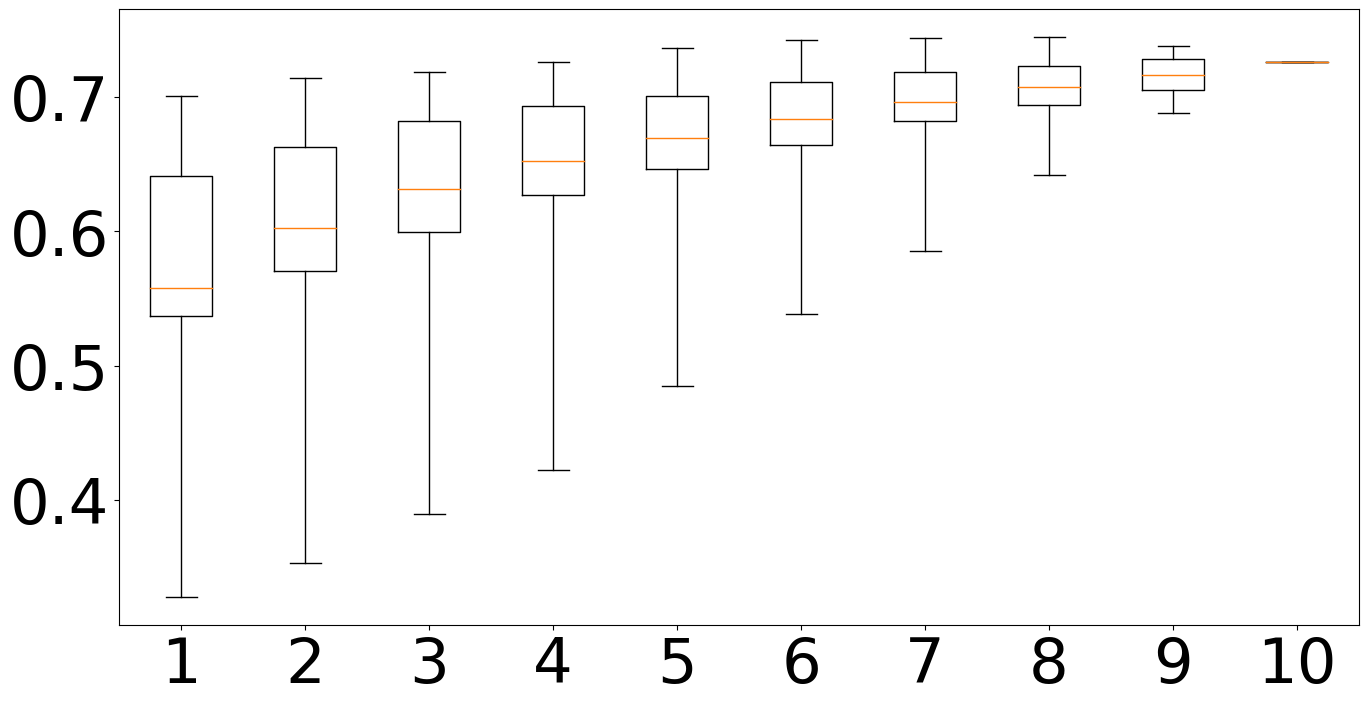} } 

\caption{
\textbf{Box plots of different numbers of criteria.}  The horizontal and vertical axes correspond to the number of criteria and the concordance index, respectively. For example, with three criteria, we select three out of ten possible criteria, resulting in 120 different combinations. The highest point, the top of the rectangle, the orange horizontal line, the bottom of the rectangle, and the lowest point respectively represent the maximum value, the top 25\%, the average value, the top 75\%, and the minimum. It is shown that as the number of criteria increases, the performance gradually improves.
}
\label{Boxplots}
\end{figure*}

\begin{table*}[!t]

    \centering

    \begin{tabular}{l|l|c|ccccc}
    \toprule
       Dataset & LLM &  Method & $i \gg j $ & $i > j $ & $i = j$ &$ j > i$ & $j \gg i  $ \\
        \midrule
         Part-time job  & ChatGPT-3.5-turbo & Pairwise  & 50.3 &  1.1 & 0  &0.1 & 48.5 \\ \hline
         Smoking  & ChatGPT-3.5-turbo & Pairwise & 42.2 &  0.2  & 0  & 0.1 & 57.5 \\ \hline
          Meeting & ChatGPT-3.5-turbo & Pairwise  & 48.7 &  0.3 & 0  &0.1 & 50.8 \\ \hline
           Cheat  &ChatGPT-3.5-turbo  & Pairwise & 24.5 &  0.0 & 0  & 0.1 & 75.4 \\ \midrule
             Part-time job  & GPT-4 & Pairwise & 16.7 &  25.6 & 0.9  & 31.0 & 25.7 \\ \hline
         Smoking  & GPT-4& Pairwise & 24.1 &  14.4 & 0  &14.8 & 46.7 \\ \hline
          Meeting & GPT-4 & Pairwise & 17.5 &  24.7 & 0.8  &18.9 & 38.4 \\ \hline
           Cheat  & GPT-4 & Pairwise & 18.2 &  21.5 & 1.3  &22.4 & 36.6 \\ \midrule 

              Part-time job  & ChatGPT-3.5-turbo & AHP  & 50.4 &  0.1 & 0  & 0.2 & 49.3 \\ \hline
         Smoking  & ChatGPT-3.5-turbo & AHP & 62.4 &  0.1 &  0  &  0.5 & 37.1 \\ \hline
          Meeting & ChatGPT-3.5-turbo & AHP  & 54.6 &  0.2 & 0  & 0.1 & 45.1 \\ \hline
           Cheat  &ChatGPT-3.5-turbo  & AHP & 48.7 &  1.2 & 0.0  & 0.3 & 49.8 \\ \midrule
             Part-time job  & GPT-4 & AHP & 21.5 &  14.6 & 0.3  & 29.1 & 34.4 \\ \hline
         Smoking  & GPT-4& AHP & 29.5 &  13.5 & 2.1  & 9.7 & 45.2 \\ \hline
          Meeting & GPT-4 & AHP & 17.5 &  39.9 & 0.1  & 28.6 & 13.9 \\ \hline
           Cheat  & GPT-4 & AHP & 25.4 &  14.8 & 0.7  & 21.2 & 37.8 \\
        \bottomrule
    \end{tabular}

      \caption{ \textbf{Statistics of LLM responses in AHP and pairwise evaluation.}  $i\gg j$, $i>j$, $i=j$, $j> i$, and $j\gg i$ correspond to the five options provided to the LLMs.}
    
\label{stat}
\end{table*}

\subsection{RQ2: What is the impact of multiple criteria on the results? -- Multiple criteria are helpful. }

Figure~\ref{Boxplots} illustrates the impact of multiple criteria. As the number of criteria increases, the average performance progressively improves. The leftmost of Figure~\ref{Boxplots} represents the performance when only one criterion is selected. We can observe a significant performance gap between the best and the worst criteria. The worst criteria often perform worse than direct pairwise comparisons without any criteria; while the average performance of all single criteria often performs worse than the all $10$ criteria combined in the AHP. Proper criteria are assigned higher importance and weight. Even if a good response scores low under less important criteria, it can still achieve high overall scores if it scores high on important criteria. Therefore, after combining all criteria, the performance of our proposed multiple criteria method is better than pairwise comparisons without criteria.

\subsection{RQ3: What are the findings in experiments with different LLMs? -- 
 The performance depends on the prompt. }
Tables~\ref{GPT3main1} and \ref{GPT4main} indicate that GPT-4 performs better in AHP and pairwise comparisons than ChatGPT-3.5-turbo. Table~\ref{stat} shows the responses of GPT-3.5-turbo and GPT-4 to five select options. ChatGPT-3.5-turbo rarely selects options containing `slightly' or `almost the same', whereas GPT-4 frequently chooses options with `slightly,' demonstrating better flexibility of evaluative scales. However, GPT-4 also rarely selects `almost the same,' despite a large number of responses at the same level in the dataset.

Nevertheless, GPT-4 does not exhibit stronger capabilities than ChatGPT-3.5-turbo in Scoring, Few-shot, and CERF Level evaluations significantly as shown in Tables~\ref{GPT3main1} and \ref{GPT4main}. Figures~\ref{Scoring}, \ref{Fewshot}, and \ref{CERF} show that the diversity of answers is not better than ChatGPT-3.5-turbo. For example, in the CERF Level evaluation for the Part-time jobs dataset, GPT-4 assigns almost all responses to Level 3. This suggests that GPT-4 does not have a better understanding of all prompts compared to ChatGPT-3.5-turbo.

%% file: conclusion.tex
\section{Conclusion}
In this study, we proposed an evaluation approach to open-ended responses by AHP-Powered LLM reasoning to evaluate open-ended responses based on multiple different criteria by LLMs. We conducted experiments on four datasets and with two types of LLMs, and we used quantitative indicators, concordance index, and soft concordance index, to demonstrate that our method outperforms the baselines empirically. Three takeaways can be summarized as follows based on the results. 
\begin{itemize}
\item In the absence of pairwise comparisons with different answers, LLMs perform poorly in directly evaluating open-ended responses, tending to assign mid-to-high-level scores to all responses. This leads to both good and poor responses receiving similar scores, resulting in a lack of differentiation. 
\item AHP-Powered multiple criteria perform better compared to pairwise comparison without criteria. Multiple criteria can effectively enhance the performance of LLMs. 
\item GPT-4 does not necessarily perform better than ChatGPT-3.5-turbo, and choosing the appropriate prompt method is crucial for relatively difficult tasks. 
\end{itemize}